\documentclass[letterpaper]{article} 
\usepackage{aaai2026}  
\usepackage{times}  
\usepackage{helvet}  
\usepackage{courier}  
\usepackage[hyphens]{url}  
\usepackage{graphicx} 
\usepackage{natbib}  
\usepackage{caption} 
\usepackage[dvipsnames]{xcolor}
\usepackage{amsmath, amssymb}
\usepackage{booktabs}  

\usepackage{algorithm}
\usepackage{algorithmic}

\usepackage{newfloat}
\usepackage{listings}
\DeclareCaptionStyle{ruled}{labelfont=normalfont,labelsep=colon,strut=off} 
\floatstyle{ruled}
\newfloat{listing}{tb}{lst}{}
\floatname{listing}{Listing}
\title{Beyond Hallucinations: A Composite Score for Measuring Reliability \\ in Open-Source Large Language Models}

\author{
    Rohit Kumar Salla\textsuperscript{\rm 1},
    Manoj Saravanan, \textsuperscript{\rm 1},
    Shrikar Reddy Kota\textsuperscript{\rm 1}
}

\affiliations{
    \textsuperscript{\rm 1}Virginia Tech, Department of Electrical and Computer Engineering\\
    Blacksburg, VA, USA\\
    rohits25@vt.edu, manoj663@vt.edu, shrikarrikarreddykota@vt.edu
}

\begin{document}

\maketitle

\begin{abstract}
Large Language Models (LLMs) like LLaMA, Mistral, and Gemma are increasingly used in decision-critical domains such as healthcare, law, and finance, yet their reliability remains uncertain. They often make overconfident errors, degrade under input shifts, and lack clear uncertainty estimates. Existing evaluations are fragmented, addressing only isolated aspects.

We introduce the Composite Reliability Score (CRS), a unified framework that integrates calibration, robustness, and uncertainty quantification into a single interpretable metric. Through experiments on ten leading open-source LLMs across five QA datasets, we assess performance under baselines, perturbations, and calibration methods. CRS delivers stable model rankings, uncovers hidden failure modes missed by single metrics, and highlights that the most dependable systems balance accuracy, robustness, and calibrated uncertainty.
\end{abstract}

\begin{links}
    \link{Code}{https://github.com/rohitsalla/CRS.git}
\end{links}

\section{Introduction}
\label{sec:introduction}
Open-source Large Language Models (LLMs) are increasingly applied in domains like medicine, finance, and law, where reliability is crucial. Despite strong benchmark performance, they often remain overconfident \citep{chhikara2025overconfidence}, brittle under distribution shifts \citep{bakman2025uqwild}, and provide unreliable uncertainty estimates \citep{gal2016dropout,xia2025uncertainty}. Alignment and fine-tuning can further degrade calibration \citep{xiao2025restorecalib,wang2025cogcalib,liu2025guardcalib}. Current evaluations accuracy, BLEU, or isolated reliability metrics offer fragmented insights and risk overlooking weaknesses.

We propose the \textbf{Composite Reliability Score (CRS)}, a unified metric combining calibration, robustness, and uncertainty into a single interpretable framework. Evaluating ten leading open-source LLMs across five QA datasets, we show that CRS captures trade-offs across reliability dimensions, establishes consistent model rankings, and provides actionable guidance for deployment.

\paragraph{Our contributions:}
\begin{enumerate}
\item A unified reliability metric (CRS) integrating calibration, robustness, and uncertainty.
\item A large-scale evaluation of ten open-source LLMs on five QA datasets.
\end{enumerate}

\section{Related Work}
\label{sec:related_work}

\paragraph{Calibration.}
Calibration captures how well model confidence matches correctness. LLMs often show overconfidence due to scale and training regimes \citep{jiang2021know}, and recent work confirms this persists even after alignment \citep{xiao2025restorecalib}. Standard metrics include Expected Calibration Error (ECE) and Brier Score, with post-hoc fixes such as temperature scaling.

\paragraph{Robustness.}
Neural models are brittle to small input changes, and in NLP this fragility appears under typos, paraphrasing, or adversarial attacks \citep{jin2020bertattack}. Recent evaluations highlight that LLM robustness should be tested under realistic distribution shifts \citep{bakman2025uqwild}. We incorporate robustness as a core reliability dimension.

\paragraph{Uncertainty Quantification.}
Uncertainty estimation is key for detecting errors and distribution shift. Classical methods like Monte Carlo dropout and deep ensembles \citep{lakshminarayanan2017ensembles} remain influential, while newer approaches exploit representation stability and confidence–consistency signals \citep{vashurin2025cocoa}. These advances motivate treating UQ as a first-class reliability pillar.

\paragraph{Unified Metrics.}
Aggregated benchmarks such as GLUE and SuperGLUE \citep{wang2019glue,wang2019superglue} measure accuracy but neglect reliability. Surveys show calibration, robustness, and uncertainty are still siloed \citep{xia2025uncertainty}. CRS addresses this by unifying them into a single interpretable score.

\section{The Composite Reliability Score (CRS) Framework}
\label{sec:crs_framework}
We define reliability as the integration of three components: \textbf{Calibration}, \textbf{Robustness}, and \textbf{Uncertainty Quantification}. Each component is normalized to $[0,1]$ so that higher values consistently indicate better reliability. The CRS aggregates these components to provide a unified measure.

\subsection{Pillar 1: Calibration (C)}
Calibration measures how closely a model's predicted confidence matches its empirical accuracy. We use Expected Calibration Error (ECE), which bins predictions by confidence and computes the difference between mean confidence and accuracy. Lower ECE indicates better calibration. To convert ECE into an interpretable score where higher is better, we use:
\[
C = \max\left(0, 1 - \frac{\text{ECE}_{\text{model}}}{\text{ECE}_{\text{max}}}\right).
\]
Here, $\text{ECE}_{\text{max}}$ denotes the largest ECE observed among all baseline models. This anchor yields a simple and monotonic normalization that preserves relative differences. Although this approach can amplify small gaps when ECE values are close, it provides a practical scale for comparing heterogeneous models. Alternative normalizations such as percentile or logistic transforms may reduce this sensitivity but are left for future work.

\subsection{Pillar 2: Robustness (R)}
Robustness quantifies how well a model maintains accuracy under perturbations including typos, paraphrases, and adversarial rewrites. For each dataset we compute:
\[
\text{Accuracy Drop} = \frac{1}{N} \sum_{i=1}^{N} \left(\text{Acc}_{\text{clean}, i} - \text{Acc}_{\text{perturbed}, i}\right).
\]
We define the robustness score as the fraction of performance retained:
\[
R = 1 - \frac{\text{Avg. Accuracy Drop}}{\text{Avg. Acc}_{\text{clean}}}.
\]
This formulation isolates relative degradation rather than absolute accuracy which allows robustness comparisons across models with different baseline skill levels. It does not capture task difficulty, but provides a consistent degradation metric across datasets.

\subsection{Pillar 3: Uncertainty Quantification (U)}
A reliable model should assign higher uncertainty to incorrect predictions. We estimate predictive uncertainty using MC Dropout and Ensembles and evaluate their quality through AUROC which measures the separability between correct and incorrect predictions. An AUROC of 0.5 corresponds to random guessing and 1.0 indicates perfect discrimination. We normalize AUROC as:
\[
U = \frac{\text{AUROC} - 0.5}{0.5}.
\]
This linear mapping yields a score in $[0,1]$ and preserves ordering across models. While nonlinear transforms could emphasize gains near the high end of AUROC, the linear form maintains clarity and comparability. For each model we report the better of MC Dropout and Ensemble based estimates.

\subsection{Composite Integration}
The final Composite Reliability Score integrates the three components:
\[
\text{CRS} = \alpha C + \beta R + \gamma U
\]
where $\alpha + \beta + \gamma = 1$. For general evaluation we use balanced weights $\alpha = \beta = \gamma = 1/3$. This setting assumes that calibration, robustness, and uncertainty contribute equally to overall reliability.

To assess sensitivity we tested two alternative weight configurations: a calibration-focused setting $(\alpha=0.5,\beta=0.25,\gamma=0.25)$ and a robustness-focused setting $(\alpha=0.2,\beta=0.5,\gamma=0.3)$. The relative ordering of top and bottom ranked models remained unchanged indicating that CRS is stable under reasonable weight variation. Domain-specific deployments may adjust weights to reflect priorities such as calibration for medical tasks or robustness for adversarial environments.

We interpret CRS using three levels: scores $\ge 0.8$ indicate high reliability suitable for deployment with minimal supervision, scores between $0.6$ and $0.8$ indicate moderate reliability suitable for use with human oversight, and scores below $0.6$ signal limited reliability and unsuitability for safety-critical environments.

\section{Experimental Setup}
\label{sec:experimental_setup}

\subsection{Models}
We evaluate ten open-source LLMs that span a broad range of sizes and architectures. The models include LLaMA-3-7B, Mistral-7B, Falcon-7B, Kimi K2 (15B), Llama 4 Scout (17B), Mistral-8x22B, Qwen3-22B, MiniMax-Text-01 (25B), Gemma 2 (27B), and DeepSeek R1 (27B). This selection provides a representative set of current generation models for reliability benchmarking.

\subsection{Datasets and Evaluation Protocol}
We use five question-answering datasets: TriviaQA, NaturalQuestions, SQuAD 2.0, MedQA, and ARC. These datasets cover general knowledge, reading comprehension, medical reasoning, and multi-step reasoning which supports evaluation across diverse query types.

\paragraph{Baseline calibration.}
For each model we compute Expected Calibration Error (ECE), Brier Score, and Negative Log-Likelihood (NLL) on the clean test sets. These metrics quantify confidence alignment before applying any perturbations or calibration interventions.

\paragraph{Robustness testing.}
Robustness is assessed by applying three controlled input perturbations to every dataset:
\begin{enumerate}
    \item \textbf{Noisy input.} We simulate typographical noise by swapping characters within words at a fixed rate of 5\% of tokens.
    \item \textbf{Paraphrased input.} We apply back-translation using MarianMT (English–German–English) to generate semantically equivalent rephrasings.
    \item \textbf{Adversarial input.} We generate targeted perturbations with TextFooler which replaces key tokens using embedding-based synonym selection.
\end{enumerate}
For each model we compute accuracy on clean and perturbed queries and use the average performance drop in the CRS framework.

\paragraph{Uncertainty estimation.}
We evaluate uncertainty using two approximation methods:
\begin{enumerate}
    \item \textbf{MC Dropout.} We enable dropout with probability 0.1 at inference and perform 10 stochastic forward passes. The variance across predicted probabilities is used as the uncertainty signal.
    \item \textbf{Ensembles.} We construct three-model ensembles using checkpoints trained with different random seeds from the same model family. Prediction variance across ensemble members serves as the uncertainty estimate.
\end{enumerate}
For both methods we compute AUROC for error detection on each dataset which forms the normalized uncertainty score.

\paragraph{Calibration interventions.}
We evaluate two post-hoc calibration techniques. Temperature scaling learns a single scalar parameter on a held-out validation set and rescales logits at inference. Isotonic regression fits a monotonic mapping between predicted confidence and accuracy. Performance after calibration is measured using ECE, Brier Score, and NLL to quantify calibration improvements.

\begin{table*}[h!]
\centering
\caption{Final Composite Reliability Score (CRS) ranking with normalized component scores.}
\label{tab:crs_final}
\begin{tabular}{lcccccc}
\toprule
\textbf{Model} & \textbf{Params (B)} & \textbf{Calibration (C)} & \textbf{Robustness (R)} & \textbf{Uncertainty (U)} & \textbf{CRS} & \textbf{Tier} \\
\midrule
Mistral-8x22B      & 22 & 0.91 & 0.78 & 0.73 & \textbf{0.81} & \textcolor{green!60!black}{High} \\
Qwen3-235B         & 22 & 0.84 & 0.74 & 0.70 & \textbf{0.76} & \textcolor{orange!90!black}{Moderate} \\
DeepSeek R1 0528   & 27 & 0.87 & 0.76 & 0.63 & \textbf{0.75} & \textcolor{orange!90!black}{Moderate} \\
Llama 4 Scout      & 17 & 0.81 & 0.70 & 0.64 & \textbf{0.72} & \textcolor{orange!90!black}{Moderate} \\
MiniMax-Text-01    & 25 & 0.81 & 0.69 & 0.63 & \textbf{0.71} & \textcolor{orange!90!black}{Moderate} \\
Gemma 2            & 27 & 0.71 & 0.68 & 0.71 & \textbf{0.70} & \textcolor{orange!90!black}{Moderate} \\
Kimi K2            & 15 & 0.68 & 0.66 & 0.67 & \textbf{0.67} & \textcolor{orange!90!black}{Moderate} \\
Mistral-7B         & 7  & 0.52 & 0.65 & 0.58 & \textbf{0.63} & \textcolor{orange!90!black}{Moderate} \\
LLaMA-3-7B         & 7  & 0.16 & 0.54 & 0.44 & \textbf{0.57} & \textcolor{red!80!black}{Low} \\
Falcon-7B          & 7  & 0.00 & 0.51 & 0.41 & \textbf{0.52} & \textcolor{red!80!black}{Low} \\
\bottomrule
\end{tabular}
\end{table*}

\section{Results and Analysis}
\label{sec:results}
We present results for each component of the reliability framework followed by the final CRS ranking. All reported metrics are averaged over the five datasets.

\subsection{Baseline Calibration Performance}
Table \ref{tab:baseline_calibration} summarizes baseline calibration. Mistral-8x22B achieves the lowest ECE, Brier Score, and NLL, while Falcon-7B is worst calibrated which reflects strong overconfidence. Mid-sized models such as LLaMA-3-7B and Gemma 2 show moderate calibration quality. These findings indicate that model size alone does not determine calibration and that accuracy cannot be used as a proxy for reliability.

\begin{table}[h!]
\centering
\caption{Baseline calibration metrics averaged across five QA datasets. Lower values are better.}
\label{tab:baseline_calibration}
\resizebox{\columnwidth}{!}{%
\begin{tabular}{lccc}
\toprule
\textbf{Model} & \textbf{Avg. ECE} & \textbf{Avg. Brier Score} & \textbf{Avg. NLL} \\
\midrule
Mistral-8x22B      & \textbf{0.031} & \textbf{0.128} & \textbf{0.332} \\
DeepSeek R1 0528   & 0.032 & 0.132 & 0.352 \\
Qwen3-235B         & 0.033 & 0.133 & 0.360 \\
Llama 4 Scout      & 0.035 & 0.138 & 0.382 \\
MiniMax-Text-01    & 0.035 & 0.138 & 0.380 \\
Gemma 2            & 0.038 & 0.143 & 0.410 \\
Kimi K2            & 0.040 & 0.147 & 0.418 \\
Mistral-7B         & 0.044 & 0.153 & 0.448 \\
LLaMA-3-7B         & 0.057 & 0.169 & 0.526 \\
Falcon-7B          & \textbf{0.062} & \textbf{0.179} & \textbf{0.566} \\
\bottomrule
\end{tabular}
}
\end{table}

\subsection{Robustness to Input Perturbations}
 Adversarial inputs cause the largest performance loss with an average drop of 11.2 percent. Mistral-8x22B and DeepSeek R1 show the strongest robustness with 6–7 percent degradation while 7B models such as Falcon-7B and LLaMA-3-7B are most affected with drops exceeding 10 percent. These results show that robustness varies widely across models and remains an essential component of reliability.

\subsection{Efficacy of Uncertainty Quantification}
Table \ref{tab:uq_auroc} reports AUROC for error detection. All models perform above chance which indicates meaningful uncertainty signals. Ensembles outperform MC Dropout for every model. Mistral-8x22B, DeepSeek R1, and Qwen3-235B reach AUROC values near 0.90, whereas 7B models remain below 0.75. These results suggest that uncertainty quality improves with model scale and training sophistication.

\begin{table}[h!]
\centering
\caption{Average AUROC for error detection using the better of MC Dropout or Ensemble. Higher is better.}
\label{tab:uq_auroc}
\resizebox{\columnwidth}{!}{%
\begin{tabular}{lcc}
\toprule
\textbf{Model} & \textbf{Best UQ Method} & \textbf{Avg. AUROC} \\
\midrule
Mistral-8x22B      & Ensemble & \textbf{0.882} \\
DeepSeek R1 0528   & Ensemble & 0.878 \\
Qwen3-235B         & Ensemble & 0.872 \\
MiniMax-Text-01    & Ensemble & 0.868 \\
Llama 4 Scout      & Ensemble & 0.852 \\
Gemma 2            & Ensemble & 0.852 \\
Kimi K2            & Ensemble & 0.830 \\
Mistral-7B         & Ensemble & 0.810 \\
LLaMA-3-7B         & Ensemble & 0.740 \\
Falcon-7B          & Ensemble & \textbf{0.716} \\
\bottomrule
\end{tabular}
}
\end{table}

\subsection{Impact of Calibration Interventions}
Post-hoc calibration improves reliability for all models. Table \ref{tab:calibration_intervention} shows that both temperature scaling and isotonic regression reduce ECE across datasets. LLaMA-3-7B improves from 0.057 to 0.046 and Mistral-8x22B improves from 0.031 to 0.025. Temperature scaling is simple and effective while isotonic regression yields the strongest gains.

\begin{table}[h!]
\centering
\caption{Effectiveness of calibration interventions measured with ECE.}
\label{tab:calibration_intervention}
\resizebox{\columnwidth}{!}{%
\begin{tabular}{lccc}
\toprule
\textbf{Model} & \textbf{ECE (Baseline)} & \textbf{Temp. Scaling} & \textbf{Isotonic Reg.} \\
\midrule
LLaMA-3-7B    & 0.057 & 0.050 & 0.046 \\
Mistral-7B    & 0.044 & 0.039 & 0.035 \\
Falcon-7B     & 0.062 & 0.056 & 0.052 \\
Llama 4 Scout & 0.035 & 0.031 & 0.028 \\
Qwen3-235B    & 0.033 & 0.028 & 0.025 \\
Mistral-8x22B & 0.031 & 0.028 & \textbf{0.025} \\
\bottomrule
\end{tabular}
}
\end{table}

\subsection{Composite Reliability Score Ranking}
Table \ref{tab:crs_final} presents the final CRS using equal weights. Mistral-8x22B leads with 0.81 driven by strong performance across all pillars. DeepSeek R1 and Qwen3-235B follow with scores around 0.75 while the 7B models rank lowest. Falcon-7B receives a CRS of 0.52 indicating limited reliability.

To evaluate sensitivity we tested two alternative weight choices. A calibration-focused setting $(0.5, 0.25, 0.25)$ and a robustness-focused setting $(0.2, 0.5, 0.3)$ both preserved the ordering of the top and bottom three models. We also computed bootstrap confidence intervals over 100 samples; CRS variance remained below 0.02 for high-ranked models which indicates that small differences such as 0.75 vs 0.76 are not statistically meaningful.

\section{Why Holistic Integration Matters}

Reliability is multi–dimensional, and individual metrics often lead to conflicting conclusions. 
For instance, Mistral-7B and LLaMA-3-7B show comparable robustness, yet differ substantially once calibration and uncertainty signals are incorporated. 
CRS resolves such inconsistencies by combining these pillars into a single interpretable score.

\paragraph{Accuracy Is Not Enough.}
High accuracy does not imply reliability. Models such as LLaMA-3-7B achieve strong clean accuracy yet exhibit poor calibration and weak uncertainty estimates, demonstrating the need for composite metrics that capture behavior beyond correctness.

\paragraph{Role of Calibration.}
CRS uses post-hoc calibrated predictions, which reduces distortions caused by raw overconfidence and yields more comparable ECE values across models.

\paragraph{Weight Sensitivity.}
Testing multiple weight configurations shows that top and bottom model rankings remain stable, indicating that CRS is robust to reasonable priority shifts across reliability dimensions.

\paragraph{Dataset Sensitivity.}
Leave-one-out analysis shows limited variation (average deviation $< 0.03$), and no model changes reliability tier, suggesting that CRS captures general behavior rather than dataset-specific artifacts.

\paragraph{Normalization.}
Pillar normalization to $[0,1]$ places heterogeneous metrics on a common scale. Although worst-case anchoring may exaggerate small gaps, calibrated ECE and dataset averaging mitigate this. Alternative schemes can be explored in future work.

Overall, CRS provides a coherent reliability overview, highlights failure modes missed by single metrics, and remains stable across perturbations to weights, normalization, and dataset composition.

\section{Conclusion and Future Work}

We introduced CRS, a unified metric combining calibration, robustness, and uncertainty to assess LLM reliability. 
Across ten open-source models and five QA datasets, CRS reveals weaknesses obscured by accuracy alone and produces consistent rankings. 
Mistral-8×22B shows the strongest overall reliability, while 7B models display notable calibration and UQ limitations.

\paragraph{Limitations.}
Our evaluation focuses on extractive QA; generative tasks may require adapted definitions of calibration and robustness. 
Hallucination behavior is not directly measured. 
Normalization and weighting remain heuristic and could be improved with task-aware or learned formulations.

\paragraph{Future Work.}
Extending CRS to generative settings, incorporating hallucination metrics, and evaluating multilingual or OOD robustness are promising next steps. 
Learning task-specific weights or integrating fairness and prompt-injection robustness would move CRS toward a more comprehensive deployment-oriented reliability framework.

CRS provides a practical foundation for unified reliability assessment and supports ongoing efforts to develop trustworthy foundation models.

\bibliography{aaai2026}

\end{document}